\documentclass{article}
\usepackage{ijcai24}
\usepackage[inkscapeformat=png]{svg}
\usepackage{times}
\usepackage{soul}
\usepackage{url}
\usepackage[hidelinks]{hyperref}
\usepackage[utf8]{inputenc}
\usepackage[small]{caption}
\usepackage{graphicx}
\usepackage{amsmath}
\usepackage{amsthm}
\usepackage{booktabs}
\usepackage{algorithm}
\usepackage{algorithmic}
\usepackage[switch]{lineno}

\usepackage{amsmath,amsfonts,bm}

\def\eqref#1{equation~\ref{#1}}

\def\1{\bm{1}}

\DeclareMathAlphabet{\mathsfit}{\encodingdefault}{\sfdefault}{m}{sl}
\SetMathAlphabet{\mathsfit}{bold}{\encodingdefault}{\sfdefault}{bx}{n}

\usepackage{tabularx}
\usepackage{hyperref}
\usepackage{url}
\usepackage{amssymb}
\usepackage{amsfonts}
\usepackage{amsmath}
\usepackage{multirow}
\usepackage{graphicx}
\usepackage{enumitem}
\usepackage{textcomp}
\usepackage{subcaption}
\usepackage{xspace} 
\usepackage{diagbox}
\usepackage{amssymb} 
\usepackage{multirow}
\usepackage{xcolor}
\usepackage{xspace} 
\usepackage{color} 
\usepackage{bm}
\usepackage{colortbl}
\usepackage{setspace}
\usepackage{array}
\usepackage{arydshln} 
\usepackage{makecell}
\usepackage{xcolor,colortbl}
\usepackage{tcolorbox}
\usepackage{listings}
\usepackage{pythonhighlight}
\usepackage{wrapfig,lipsum,booktabs}
\usepackage{hyperref}

\newcommand{\taco}{\textsc{TACO}\xspace}

\makeatletter
\renewcommand*{\@fnsymbol}[1]{%
  \ensuremath{%
    \ifcase#1\or \dagger\or *\or \ddagger\or
    \mathsection\or \mathparagraph\or \|\or **\or \dagger\dagger
    \or \ddagger\ddagger \else\@ctrerr\fi}}
\makeatother

\title{\includegraphics[height=1.2\fontcharht\font`A]{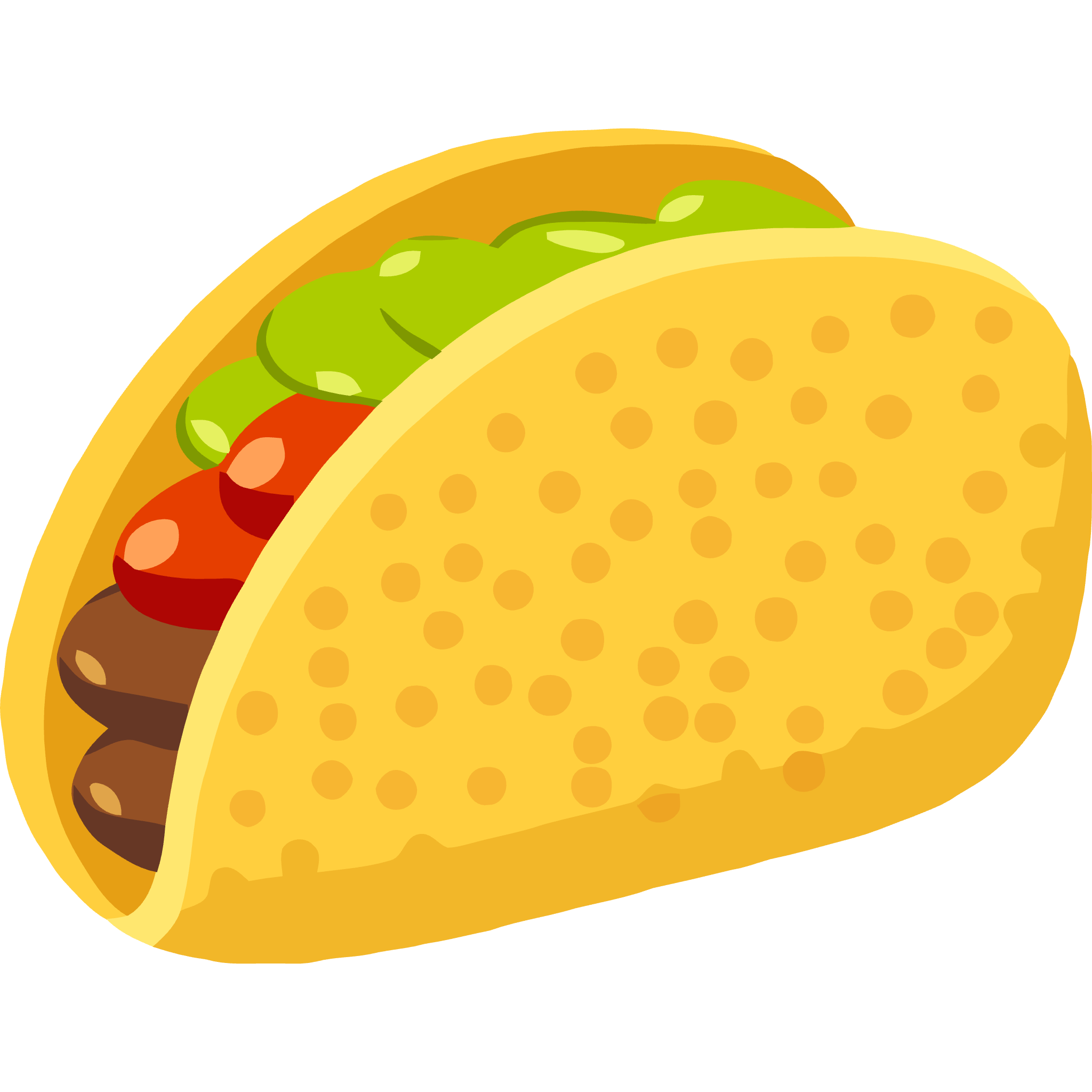} \taco: Topics in Algorithmic COde generation dataset}

\author{
Rongao Li$^{1,}$\thanks{Equal technical contribution.}\and
Jie Fu$^{1,\dagger}$\and
Bo-Wen Zhang$^{1,\dagger,}$\thanks{Corresponding Authors}\and
Tao Huang$^{2}$\and
Zhihong Sun$^{2}$\and\\
Chen Lyu$^{2,*}$\and
Guang Liu$^{1}$\and
Zhi Jin$^{3}$\And
Ge Li$^{3}$
\affiliations
$^1$Beijing Academy of Artificial Intelligence\\
$^2$School of Information Science and Engineering, Shandong Normal University, China\\
$^3$Key Lab of HCST (PKU), MOE; SCS, Peking University, China\\
\emails
bwzhang@baai.ac.cn, lvchen@sdnu.edu.cn
}

\begin{document}

\maketitle

\begin{abstract}
We introduce TACO, an open-source, large-scale code generation dataset, with a focus on the topics of algorithms, designed to provide a more challenging training dataset and evaluation benchmark in the field of code generation models. TACO includes competition-level programming questions that are more challenging, to enhance or evaluate problem understanding and reasoning abilities in real-world programming scenarios. There are 25433 and 1000 coding problems in training and test set, as well as up to 1.55 million diverse solution answers. Moreover, each TACO problem includes several fine-grained labels such as task topics, algorithms, programming skills, and difficulty levels, providing a more precise reference for the training and evaluation of code generation models. The dataset and evaluation scripts are available on \href{https://huggingface.co/datasets/BAAI/TACO}{Hugging Face Hub}\footnote{https://huggingface.co/datasets/BAAI/TACO} and \href{https://github.com/FlagOpen/TACO}{Github}\footnote{https://github.com/FlagOpen/TACO}.
\end{abstract}
\section{Introduction}
Code capability is one of the core competencies of foundational models, crucial for enhancing key skills such as inference and planning in these models. In recent years, large-scale language models (LLMs) based on the Transformer architecture \cite{vaswani2017attention,brown2020language} have made significant advancements in applications involving generating code from a natural language description, establishing themselves as the foundational backbone for a myriad of code-related downstream tasks \cite{chen2021evaluating,nijkamp2022codegen,li2023starcoder,Rozire2023CodeLO}. By treating the natural language to code as a sequential transformation task, these models offer solutions to basic programming problems \cite{austin2021program}. Nevertheless, they encounter substantial difficulties when addressing complex and novel programming challenges \cite{hendrycks2021measuring,li2022competition,liu2023your}. Recent evaluations indicate that the GPT-4 model performs remarkably poorly in programming challenges when compared to other human-level benchmarks, achieving only a 7\% success rate in Leetcode (hard) tasks \cite{OpenAI2023GPT4TR}.

With the rapid development of large language models and code generation models, mainstream code evaluation benchmarks have begun to reveal their limitations, struggling to comprehensively reflect the models' performance and potential in real-world scenarios.

\begin{itemize}
    \item Low Difficulty of Evaluation Tasks: The current mainstream benchmarks like HumanEval/HumanEval-X, MBPP/MBXP primarily focus on basic programming problems, requiring the model to complete a function or a Text2Code task, rather than solving a real-world problem. Moreover, the performance scores on these benchmarks are already high, with HumanEval's state-of-the-art (SOTA) model achieving 94.4~(pass@1) and MBPP's SOTA model at 81.1~(Acc). Consequently, the reference value of these evaluation results is gradually decreasing.
    \item Test Set Quality Issues, Doubts on Validity: Code execution and test cases are key to verifying the correctness of code. However, benchmarks like APPS and CodeContest still face issues with their test sets, such as the absence of manual answers and non-de-duplicated problem and answer sets. Additionally, a DeepMind paper titled 'Competition-Level Code Generation with AlphaCode' pointed out that due to insufficient test cases, code evaluation datasets might encounter the issue of False Positives, where all test cases pass, but a manual review reveals incorrect code implementation, indicating the model only just passed the given few test cases.
    \item Lack of Fine-Grained Indicators: Current code evaluation datasets lack more fine-grained indicators for assessing coding ability, such as difficulty dimensions, algorithm dimensions (sorting, dynamic programming, etc.), failing to provide targeted guidance for improving model code generation capabilities.   
\end{itemize}

Therefore, we need a more challenging, higher quality, and more finely-grained code generation evaluation scheme to assess models' code generation capabilities and to provide guidance for enhancing these capabilities.

Programming tasks usually exhibit a pronounced heterogeneity; they possess disparate requirements, necessitate various problem-solving strategies, and employ diverse algorithms and data structures. In this paper, we propose a novel perspective by modeling program synthesis as a combination of multiple tasks~\cite{caruana1997multitask}, treating different programming challenges as instance-level tasks. Therefore, the foundation is the detailed annotation of code data with respect to algorithm topics. Existing code datasets either lack comprehensive annotations or provide them in an excessively generalized form. However, algorithms are like the basic alphabet of programming, influencing every coding decision and laying the groundwork for higher-level abstractions. We suggest that enhanced algorithmic annotations have the potential to improve progress in the domain of code understanding and generation, thereby offering benefits to the community. To this end, we have curated and publicly released a large code dataset named \taco(\emph{\textbf{T}opics in \textbf{A}lgorithm for \textbf{Co}de}), which is collected from a variety of sources. \taco consists of 26,443 programming tasks, spanning a wide range of subjects including mathematics, data structures, and graph theory. Notably, \taco offers a rich correlation between programming tasks and their respective algorithms, and also exhibits richness in annotations related to algorithms, including aspects like time complexity. Drawing upon the insights offered by \cite{laaksonen2017competitive}, we have meticulously revised and systematically categorized raw tags from the original datasets into 36 primary topics in algorithm, including \emph{Complete Search} and \emph{Dynamic Programming}. Additionally, we have supplemented these categories with annotations denoting foundational programming techniques, herein referred to as skills. This aims to assist models in mastering the essential competencies required for solving wide range of programming challenges. Given its comprehensive scope and diverse set of tasks, \taco serves as a valuable benchmark for evaluating the model's capability to address a wide array of programming challenges across various algorithmic topics.

\section{Overview and Characteristics}
\label{sec:appendix_detail}
TACO dataset comprises problems sourced from publicly accessible datasets and a curated set of manually verified open-source problems. While C++ predominates in programming competitions, our data collection efforts centered on Python 3 solutions. This decision is underpinned by Python's more restrictive syntactic rules relative to C++, leading to cleaner and more discernible code structures. The TACO dataset encompasses not only algorithmic competencies but also multi-faceted metadata including time and space constraints, timestamps, among others, representing its most salient point of differentiation from extant datasets.

Following an exhaustive examination, we identified multiple shortcomings in existing open-source code generation datasets. These limitations include a restricted range of problems and an absence of metadata that elucidates the algorithmic skills required to solve the tasks. Within the realm of programming challenges, problems are often addressed by employing a confined yet essential repertoire of strategies, encompassing algorithms like dynamic programming, exhaustive search, and sorting techniques. Generally speaking, solutions to these challenges are an amalgamation of established methodologies and innovative insights.

To facilitate the more effective incorporation of algorithmic competencies into code generation tasks, we have developed the TACO dataset. This dataset offers the following distinct contributions and advantages in comparison to existing open-source code generation datasets:

\begin{figure}[t]
  \centering
  \includegraphics[width=\linewidth]{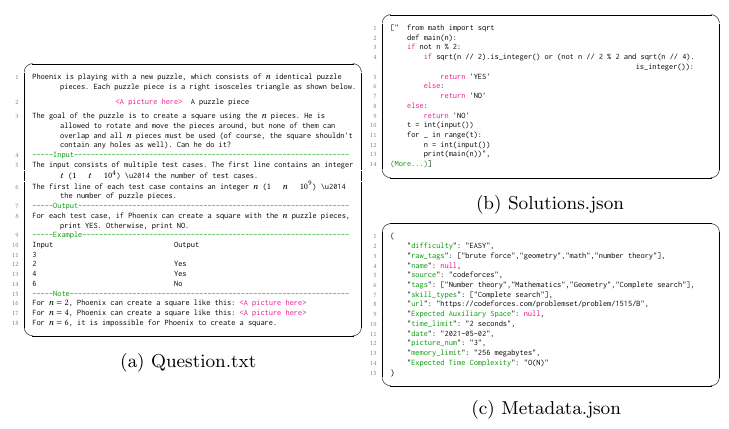}
  \caption{An example of a classical algorithmic problem (e.g., quertion, solutions, and lables)}
  \label{fig:1}
\end{figure}

\begin{itemize}
\item {\textbf{More Extensive:}} The TACO dataset comprises 26,443 problems (The training set has 25,443 problems and the test set has 1,000 problems) and 1,539,152 verified Python 3 code solutions. Initially, we amassed 2,045,502 code examples, but this number was reduced following a rigorous code deduplication process. On average, each problem features 58.21 correct code solutions.(Each problem in the training set has 51.6 test cases, and the test set has 202.3.)
\item {\textbf{Algorithmic Categorization:}} The TACO dataset emphasizes the algorithmic labeling of problems encountered in programming challenges. Across all problems, we thoroughly summarized the algorithmic labels of the 968 categories, ultimately consolidating them into 36 distinct algorithmic categories.
\item {\textbf{Data Quality Verification:}} We have performed a manual correctness assessment on a subset of data collected from the TACO dataset. To ensure the accuracy of test cases and solutions within the dataset, we have carried out comprehensive verification using unit tests.
\item {\textbf{Performance Benchmarks:}} The benchmarks we propose primarily serve to assess the code-generation efficacy of the nl2code model. Within the test set, we offer an exhaustive coverage of disparate algorithms and problems, aiming to gauge the model's performance across a spectrum of algorithmic skills pivotal to problem-solving.
\end{itemize}

An illustrative data sample is provided in Figure~\ref{fig:1}. 


\section{Dataset Construction}
The development of the TACO dataset is meticulously divided into two distinct phases: data collection and data processing. In the data collection phase, we focused on collecting raw data primarily from a diverse range of programming competition platforms. This phase was further enriched by integrating additional open-source datasets, thereby expanding the dataset's scope and variety. Following the collection, the data processing phase began, involving a series of intricate steps. Initially, this phase entailed code decommenting and deduplication to ensure data quality and uniqueness. Subsequently, we augmented the test samples to enhance the robustness of the dataset. In this phase, we also engaged in the classification and generalization of tasks and skill sets, using insights gained from a variety of programming contest websites and open source code generation datasets. This comprehensive approach allowed us to establish a well-defined and detailed format schema for the TACO data set.

\subsection{Data collection phase}
During the assembly of the TACO dataset, we conducted an exhaustive review of various programming contest platforms as well as existing code generation datasets. Due to legal restrictions and the intricate nature of platform-specific API interfaces, our primary data sources were platforms such as CodeChef, CodeForces, HackerRank, and GeeksforGeeks. We also integrated existing datasets, including APPS, CodeContest, and Description2code. To construct and optimize the TACO dataset, we employed two primary methodologies: First, we utilized web scraping techniques to extract problem information from programming competition websites; second, we augmented existing open-source datasets with algorithmic skill labels and other relevant tags.

To guarantee the dataset's integrity and the comprehensiveness of its tagging metadata, we devised bespoke HTML parsers tailored to each of the four major programming contest platforms: CodeChef, CodeForces, HackerRank, and GeeksforGeeks. These parsers are engineered to accommodate the diverse syntactic layouts present in the problem texts of their source websites. Throughout the parser development phase, we confirmed the accuracy of each specialized parser via extensive manual sampling. On exceptional occasions, we also employed OCR tool APIs (e.g., SimplePix) to facilitate the conversion of svg-formatted information into LaTeX format. For a detailed discussion of the validation procedures and data scraping techniques used for these platforms, refer to the Appendix~\ref{appendix:implement}. Additionally, with regard to the algorithmic labeling designed to augment the APPS and CodeContest datasets, our HTML parser similarly crawls algorithmic skill labels from sources such as LeetCode, Kattis, CodeWars, and CodeForces.

\begin{figure}[t]
  \centering
  \includegraphics[width=\linewidth]{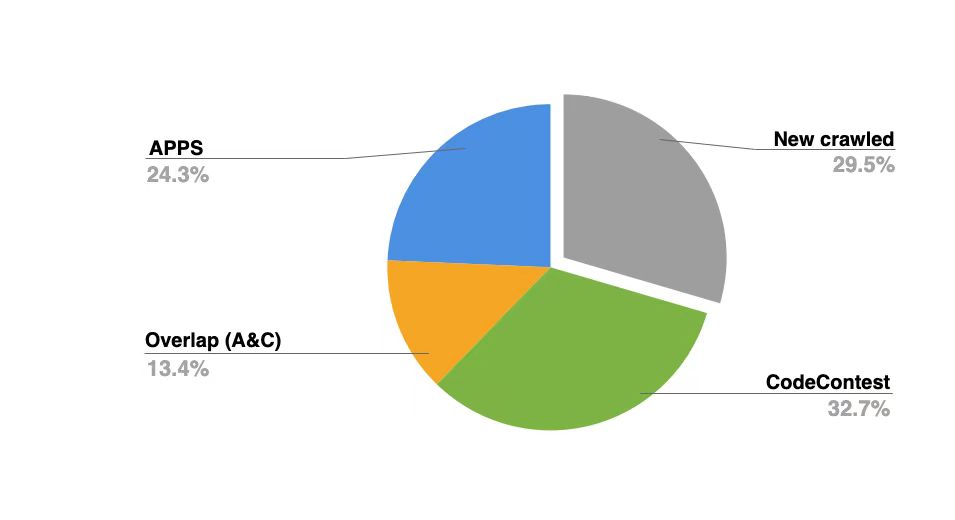}
  \caption{Analysis of TACO dataset composition}
  \label{fig:2}
\end{figure}

Upon completion of the assembly of the TACO dataset, we performed a rigorous analysis to assess the degree of overlap between TACO, APPS, and CodeContest in terms of the quantity of the problem. Initially, we executed problem correspondences predicated on the URLs corresponding to each problem. Subsequently, we expanded our correspondence framework to include plaintext analyses, which entailed the removal of whitespace and newline characters. Our calculations revealed that there are 3,392 duplicate problems shared between APPS and CodeContest. The degree of overlap in the problem between TACO and CodeContest is 0.4617, while the overlap coefficient with APPS is 0.3778. As depicted in Figure~\ref{fig:2}, 29.5\% of the problems constituting the TACO dataset are sourced from our recent web crawling endeavors; among the remaining problems, 24.3\% are exclusive to the APPS dataset, 32.7\% are unique to CodeContest, and the residual 13.4\% are featured in both APPS and CodeContest.

As detailed in Table~\ref{tab:source}, the TACO dataset includes data from multiple platform programming contests, listing the diverse data sources corresponding to each website. For websites with multiple sources of data, we have executed data merging and deduplication procedures. It is noteworthy that certain platforms, such as Aizu, AtCoder, and HackerEarth, already feature data integrated from CodeContests and Description2code, and include the requisite algorithmic tags; therefore, additional crawling efforts were deemed unnecessary for these sites.


\begin{table*}[]
\caption{Websites and access to data sources in TACO}
\label{tab:source}
\centering
\renewcommand{\arraystretch}{1.3}
\begin{tabular}{
>{\columncolor[HTML]{FFFFFF}}l |
>{\columncolor[HTML]{FFFFFF}}c |
>{\columncolor[HTML]{FFFFFF}}c |
>{\columncolor[HTML]{FFFFFF}}c }
\hline
{\color[HTML]{1F2328} \textit{\textbf{Site}}}          & {\color[HTML]{1F2328} \textit{\textbf{URL}}}         & {\color[HTML]{1F2328} \textit{\textbf{Source}}}                                                 & {\color[HTML]{1F2328} \textit{\textbf{Web Crawler Details}}}                                                                   \\ \hline
{\color[HTML]{1F2328} \textit{\textbf{Aizu}}}          & {\color[HTML]{000000} https://judge.u-aizu.ac.jp}    & {\color[HTML]{000000} CodeContests}                                                             & {\color[HTML]{000000} None}                                                                                                    \\ \hline
{\color[HTML]{1F2328} \textit{\textbf{AtCoder}}}       & {\color[HTML]{000000} https://atcoder.jp}            & {\color[HTML]{000000} \begin{tabular}[c]{@{}c@{}}APPS\\ CodeContests\end{tabular}}              & {\color[HTML]{000000} None}                                                                                                    \\ \hline
{\color[HTML]{1F2328} \textit{\textbf{CodeChef}}}      & {\color[HTML]{000000} https://www.codechef.com}      & {\color[HTML]{000000} \begin{tabular}[c]{@{}c@{}}CodeContests\\ CodeChef\end{tabular}}          & {\color[HTML]{000000} \begin{tabular}[c]{@{}c@{}}Problem\&Solution\\ Algo tags\\ Timestamp\\  Time, memory limit\end{tabular}} \\ \hline
{\color[HTML]{1F2328} \textit{\textbf{Codeforces}}}    & {\color[HTML]{000000} https://codeforces.com}        & {\color[HTML]{000000} \begin{tabular}[c]{@{}c@{}}APPS\\ CodeContests\\ Codeforces\end{tabular}} & {\color[HTML]{000000} \begin{tabular}[c]{@{}c@{}}Problem\&Solution\\ Algo tags\\ Timestamp\\  Time, memory limit\end{tabular}} \\ \hline
{\color[HTML]{1F2328} \textit{\textbf{CodeWars}}}      & {\color[HTML]{000000} https://www.codewars.com}      & {\color[HTML]{000000} APPS}                                                                     & {\color[HTML]{000000} Algo tags}                                                                                               \\ \hline
{\color[HTML]{1F2328} \textit{\textbf{GeeksforGeeks}}} & {\color[HTML]{000000} https://www.geeksforgeeks.org} & {\color[HTML]{000000} GeeksforGeeks}                                                            & {\color[HTML]{000000} \begin{tabular}[c]{@{}c@{}}Problem\&Solution\\ Algo tags\\ Time, space complexity\end{tabular}}          \\ \hline
{\color[HTML]{1F2328} \textit{\textbf{HackerEarth}}}   & {\color[HTML]{000000} https://www.hackerearth.com}   & {\color[HTML]{000000} Description2code}                                                         & {\color[HTML]{000000} None}                                                                                                    \\ \hline
{\color[HTML]{1F2328} \textit{\textbf{HackerRank}}}    & {\color[HTML]{000000} https://www.hackerrank.com}    & {\color[HTML]{000000} HackerRank}                                                               & {\color[HTML]{000000} \begin{tabular}[c]{@{}c@{}}Problem\&Solution\\ Algo tags\end{tabular}}                                   \\ \hline
{\color[HTML]{1F2328} \textit{\textbf{Kattis}}}        & {\color[HTML]{000000} https://open.kattis.com}       & {\color[HTML]{000000} APPS}                                                                     & {\color[HTML]{000000} Algo tags}                                                                                               \\ \hline
{\color[HTML]{1F2328} \textit{\textbf{LeetCode}}}      & {\color[HTML]{000000} https://leetcode.com}          & {\color[HTML]{000000} APPS}                                                                     & {\color[HTML]{000000} Algo tags}                                                                                               \\ \hline
\end{tabular}
\end{table*}

\subsection{Data processing phase}
While we exerted considerable effort during the data collection phase to ensure the high fidelity of acquired problem information and Python code to their respective source websites, it is crucial to acknowledge that the majority of the code found on programming competition platforms was manually authored and submitted by participants. This inherently introduces the possibility of encountering annotated or duplicated code. To mitigate this challenge, we implemented a robust series of correctness processing and functional processing on the dataset.

\subsubsection{Correctness processing}

\begin{itemize}
\item {\textbf{Unit Test Validation:}} While the code harvested from programming contest platforms was ostensibly marked as ``correct'' by those sites, our examination revealed instances where some code either was incorrect or failed to meet predefined criteria, such as time or space complexity, time or space limitations, or adherence to coding standards. To guarantee that the code within our dataset met a robust standard of correctness, we executed each solution against a comprehensive set of predefined unit tests that assessed these specific criteria, subsequently omitting any that failed to pass this rigorous evaluation.
\item {\textbf{Conversion from Python 2 to Python 3:}} Upon incorporating a set of problems from HackerEarth, sourced from the Description2code dataset, we observed that the provided solutions were implemented in Python 2, which was incongruent with our focus on Python 3 code. To rectify this discrepancy, we employed Python's 2to3 library for automated code translation. Solutions that could not be successfully converted to Python 3 were excluded, retaining only those that could be transitioned to the Python 3 format as part of the final dataset.
\end{itemize}

\subsubsection{Functional processing}
\begin{itemize}
\item {\textbf{Code De-duplication: }} Given that our data collection draws from solutions submitted by multiple users and integrates various open-source datasets, the propensity for encountering similar or duplicate code within individual problems is elevated. To mitigate this issue, we employed a targeted deduplication strategy. Specifically, we utilized the MinHash algorithm in conjunction with the Jaccard similarity index to achieve near-deduplication. A Jaccard similarity threshold was established at 0.85, and MinHash-assisted clusters of duplicate code were generated. Subsequently, these clusters were condensed into unique code files based on the Jaccard similarity metric.
\item {\textbf{Code De-commenting:}} In the code submissions retrieved from programming contest platforms, such as CodeChef, we observed a significant prevalence of comments. These comments, while ostensibly redundant, not only inflate the code's length but also introduce potential noise into subsequent analyses. To address this issue, we employed the Abstract Syntax Tree (AST) parsing methodology to remove these comments. Specifically, the collected code is initially converted to AST format, whereupon comment nodes are identified and eradicated. Following this, the AST structure is reverted to its original code format. It is noteworthy that the AST parsing technique also affords us the capability to execute additional code formatting tasks, such as the elimination of superfluous whitespace.
\item {\textbf{Supplementation of Test Cases}} Owing to the unavailability of hidden unit tests on programming websites for crawling, most of the problems within the APPS dataset are restricted to one or two sets of unit tests, as provided in the problem description. This constraint not only undermines the precise evaluation of code correctness but also risks the generation of incorrect code that may pass these limited unit tests. According to AlphaCode, the APPS dataset exhibited a False Positive (FP) Rate of up to 60\% when assessed with an average of 20 unit tests, thereby signaling a lack of stringent evaluation criteria. To rectify this limitation, we augmented the code samples with additional unit tests, elevating the average number of unit tests to over 200. AlphaCode's research indicates that the FP rate can be substantially reduced to approximately 4\% when the average number of unit tests exceeds 200. Specifically, we used OpenAI's GPT-4 API to generate the input components of unit test pairs. Subsequently, 30 verified correct code samples were executed on these inputs to generate corresponding outputs, which were then scrutinized for consistency. Input-output pairs that produced consistent outputs were considered valid unit tests. This iterative process was carried out multiple times to ensure that the total number of unit tests reached a minimum threshold of 200.
\end{itemize}

To meet our research objectives, we employed the aforementioned methodology and executed meticulous data processing on the multiple components of the TACO dataset, as delineated in Table~\ref{tab:details}. Specifically, we undertook a rigorous data-cleaning process for the newly acquired problems from CodeChef, CodeForces, HackerRank, and GeeksforGeeks. Likewise, we conducted an exhaustive cleaning procedure on problems obtained from open-source datasets—namely APPS, CodeContest, and Description2code—with the objective of eliminating any invalid information from the solutions, thereby enhancing the accuracy and reliability of the TACO dataset.

\begin{table*}[]
\centering
\caption{Details on data processing for each component of the TACO dataset}
\label{tab:details}
\renewcommand{\arraystretch}{1.3}
\begin{tabular}{l|c|c|c|c}
\hline
\textit{\textbf{Components}}                                                                      & \multicolumn{1}{l|}{\textit{\textbf{\begin{tabular}[c]{@{}l@{}}New crawled \\ programming problem\end{tabular}}}} & \multicolumn{1}{l|}{\textit{\textbf{APPS}}} & \multicolumn{1}{l|}{\textit{\textbf{CodeContest}}} & \multicolumn{1}{l}{\textit{\textbf{Description2code}}} \\ \hline
\textit{\textbf{Unit Test Validation}}                                                            & {\color[HTML]{000000} Yes}                                                                                        & {\color[HTML]{000000} No}                   & {\color[HTML]{000000} No}                          & No                                                     \\ \hline
\textit{\textbf{\begin{tabular}[c]{@{}l@{}}Conversion from \\ Python 2 to Python 3\end{tabular}}} & {\color[HTML]{000000} No}                                                                                         & {\color[HTML]{000000} No}                   & {\color[HTML]{000000} No}                          & Yes                                                    \\ \hline
\textit{\textbf{Code De-duplication}}                                                             & Yes                                                                                                               & Yes                                         & Yes                                                & Yes                                                    \\ \hline
\textit{\textbf{Code De-commenting}}                                                              & Yes                                                                                                               & Yes                                         & Yes                                                & Yes                                                    \\ \hline
\end{tabular}
\end{table*}

\begin{table*}[]
\caption{Comparative analysis of TACO with other datasets}
\label{tab:compare}
\centering
\renewcommand{\arraystretch}{1.3}
\begin{tabular}{l|c|c|c}
\hline
\textit{\textbf{Dataset}}                                                                                & \textit{\textbf{APPS}}        & \textit{\textbf{CodeContests}} & \textit{\textbf{TACO}}   \\ \hline
\textit{\textbf{Problems}}                                                                               & {\color[HTML]{000000} 10000}  & {\color[HTML]{000000} 13610}   & {\color[HTML]{000000} 26443}   \\ \hline
\textit{\textbf{Solutions}}                                                                              & {\color[HTML]{000000} 232421} & {\color[HTML]{000000} 1502532} & {\color[HTML]{000000} 1539152} \\ \hline
\textit{\textbf{Algorithmic labeling or not}}                                                            & No                            & No                             & Yes                            \\ \hline
\textit{\textbf{\begin{tabular}[c]{@{}l@{}}Problems without python \\ solution in testset\end{tabular}}} & 1235/5000(24.7\%)             & 43/165(26.1\%)                 & 0/1000(0\%)                    \\ \hline
\textit{\textbf{Proportion of valid problems}}                                                          & 87.65\%                       & 61\%                           & 78\%                           \\ \hline
\end{tabular}
\end{table*}

\subsubsection{Skill algorithm categorization }
\begin{figure}[t]
  \centering
  \includegraphics[width=\linewidth]{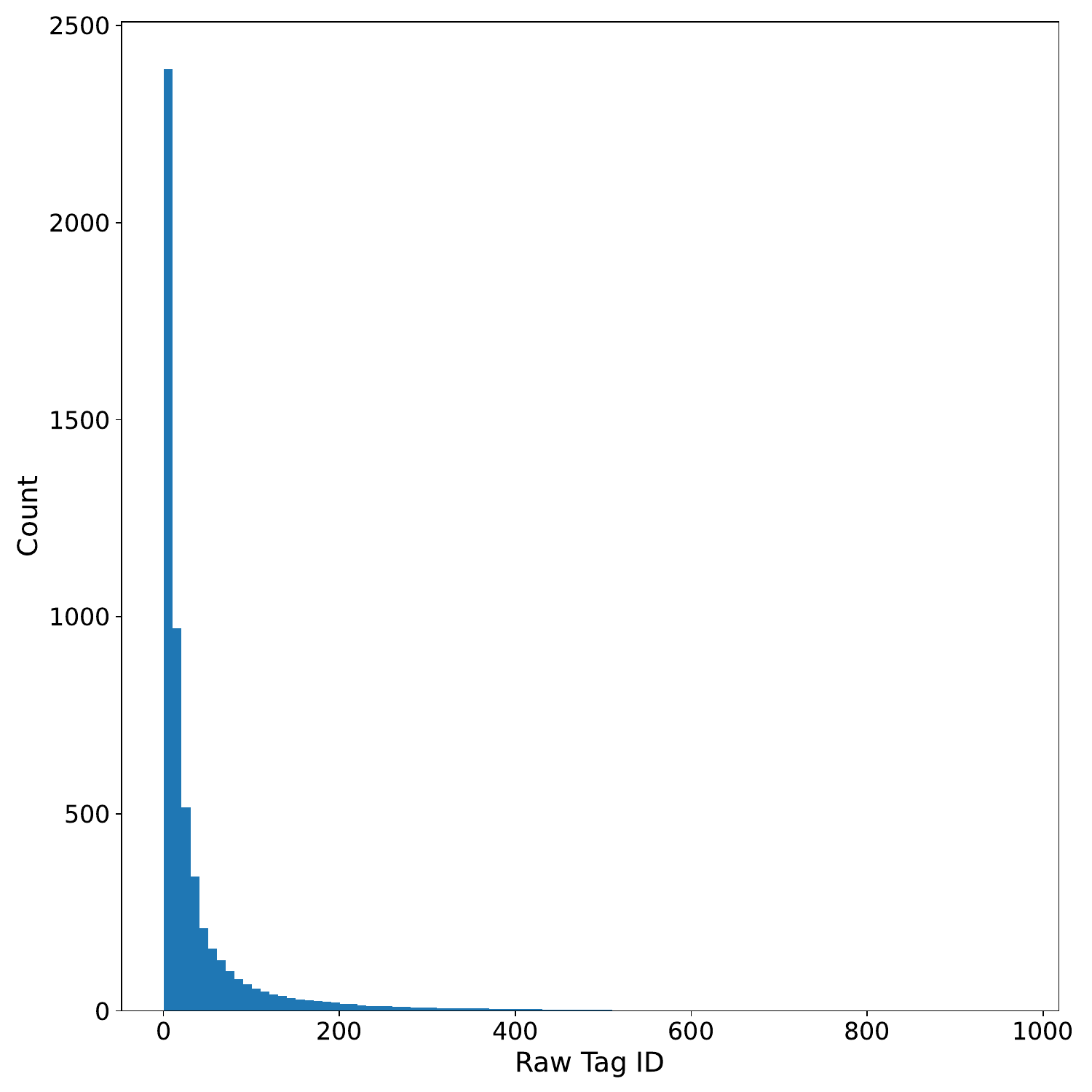}
  \caption{Statistical distribution of labels from original sources on the dataset}
  \label{fig:3}
\end{figure}

In the Source Programming Problems dataset, we retained 968 category original labels spanning various domains. These labels were manually annotated by domain experts and offer the most granular level of informational categorization. Specifically, these labels may encapsulate either the subject matter of the problem (e.g., mathematics, geometry, graphs, or strings) or feasible solution strategies (e.g., dynamic programming, brute-force approaches, or binary search). It is noteworthy that a single problem may have multiple such labels simultaneously. For a comprehensive view of the original label distribution across all problems in the dataset, refer to Figure~\ref{fig:3}. As evidenced by this figure, the distribution of these foundational labels is highly irregular and diverse. Consequently, it is crucial to systematically categorize and generalize these algorithmic labels.

A usability analysis was undertaken to elucidate both the quantity and distribution of raw labels. Our findings suggest that an excessive and intricate array of algorithmic labels could potentially impede the model's training and inference processes. Consequently, we conducted a thorough generalization of the initial set of 968 category algorithmic labels. Guided by three key considerations, we ultimately reclassified all labels into 36 discrete categories.
\begin{itemize}
    \item \textbf{Coverage: }The chosen tags must offer comprehensive encapsulation of both programming techniques and algorithmic design methods pertinent to programming challenges.
    \item \textbf{Balance Between Specificity and Generality: }Overly specific labels could impede a comprehensive grasp of programming problems, whereas excessively general labels might fall short in furnishing detailed insights into the problem. Thus, judicious abstraction and generalization of the original tags are imperative.
    \item \textbf{Label Quality:} Following meticulous organization and generalization, the resulting set of labels should accurately and profoundly capture the quintessential attributes of programming problems and their corresponding solutions.
\end{itemize}

\begin{figure}[t]
  \centering
  \includegraphics[width=\linewidth]{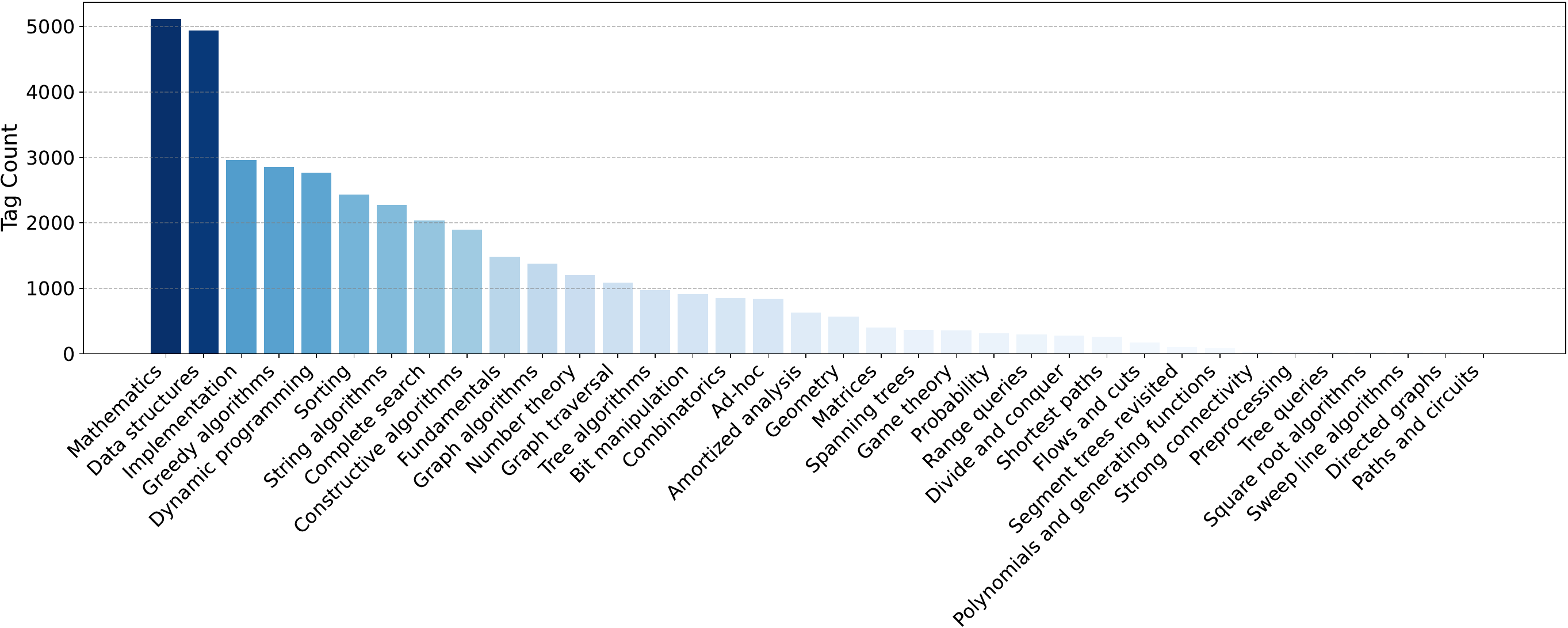}
  \caption{Statistical distribution of 36 algorithmic labels obtained by generalization on the dataset}
  \label{fig:4}
\end{figure}

\begin{figure}[t]
  \centering
  \includegraphics[width=\linewidth]{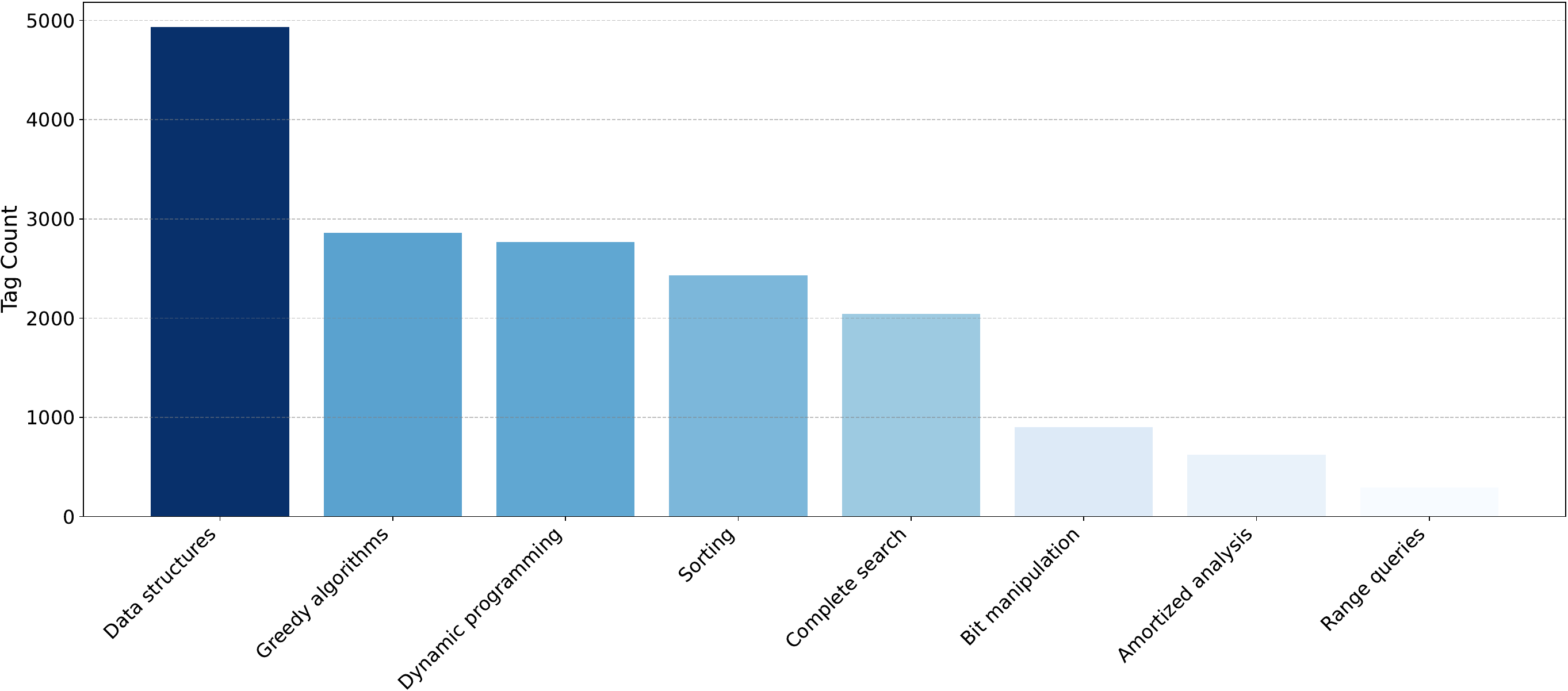}
  \caption{Statistical distribution of the 8 skill labels obtained by generalization on the dataset}
  \label{fig:5}
\end{figure}

 As illustrated in Figure~\ref{fig:4}, we have consolidated the fine-grained labels into 36 coarse-grained algorithmic labels. To enhance the adaptability of the TACO dataset across various customized application scenarios, we keep the problem's original label, algorithm label, and skill label in the dataset's metadata file. Users may consult this partial content to reclassify or amalgamate labels in accordance with their specific requirements. Moreover, we use the eight Basic Techniques enumerated in the Competitive Programmer’s Handbook as a benchmark for statistically analyzing the distribution of these foundational skills. At the same time, the 8 skill tabs are included in the 36 algorithm tabs. Quantitative information can be found in Figure~\ref{fig:5}.

\section{Comparisons with Code Datasets}
The principal code generation datasets commonly used in programming competitions include APPS and CodeContest. In the development of the TACO dataset, we utilized these two existing datasets as reference standards and implemented a series of enhancements and extensions to mitigate their limitations. Specifically, our cardinal contribution was the integration of systematic algorithmic annotations for each problem within the dataset. Concurrently, we observed that a notable proportion of problems within the APPS and CodeContest test sets lacked human-submitted Python solutions, a deficiency that hampers effective validation of model inference.

Regarding model validation, APPS principally employs GPT-2 and GPT-Neo as benchmark models, thereby neglecting an evaluation of the fine-tuning capabilities of contemporary code models, specifically in the context of programming challenges. In contrast, our TACO dataset comprises a suite of benchmarks generated through the fine-tuning of state-of-the-art models; notably, the most extensive among these employs the StarCoder model, which boasts 15 billion parameters (refer to Section \ref{sec:results} for further details).

We have counted the number of problems, the number of solutions, the presence or absence of algorithmic labels, and the number of problems without python solutions in the test set for the three datasets TACO, APPS, and CodeContests, and these differences are enumerated in detail in Table~\ref{tab:compare}. In addition, we provide more details about each dataset as follows:

\textbf{APPS: }The APPS dataset primarily originates from programming competition platforms, including CodeForces, LeetCode, and CodeChef, among others. Both the training and test sets of this dataset comprise 5,000 problems. Notably, existing academic literature, particularly the study conducted by Alphacode, highlights that the APPS dataset employs only samples drawn from problem descriptions to serve as the HIDDEN TEST CASE, leading to an elevated False Positive Rate.

\textbf{CodeContests: }The CodeContests dataset encompasses an extensive array of popular programming languages, such as C++, C\#, Go, Java, JavaScript, Lua, PHP, Python, Ruby, Rust, Scala, and TypeScript. The dataset is partitioned into training, validation, and test sets, containing 13,328, 117, and 165 problems, respectively. In contrast to datasets that exclusively feature correct code, CodeContests incorporates a variety of erroneous code types.

\textbf{TACO: }The TACO benchmark suite encompasses a wide array of algorithms programming challenges. Its primary objective is to rigorously assess a model's proficiency in mastering various algorithmic competencies integral to problem-solving. For each problem within the TACO dataset, we offer comprehensive algorithmic labels. In contrast to APPS and CodeContests, TACO supports not only the Standard Input Format (as is found in APPS) but also introduces a Call-Based Format. Furthermore, code examples are accompanied by detailed annotations, which include original labels, algorithmic tags, skill categorization, as well as time and space constraints, timestamps, and so forth.
\section{Code Generation Evaluation}

\begin{table}[]
\caption{Distribution of difficulties and programming skills in TACO training and test dataset}
\label{tab:dist}
\begin{tabular}{|c|c|c|c|}
\hline
\textbf{Category} & \textbf{Type} & \textbf{Train} & \textbf{Test} \\ \hline
\multirow{6}{*}{\textbf{Difficulty}} & \textbf{Easy} & 8904 & 200 \\
 & \textbf{Medium} & 3244 & 200 \\
 & \textbf{Medium\_Hard} & 2745 & 200 \\
 & \textbf{Hard} & 3162 & 200 \\
 & \textbf{Very\_Hard} & 2374 & 200 \\
 & \textbf{Unknown} & 5014 & 0 \\ \hline
\multirow{9}{*}{\textbf{Skill}} & \textbf{Range Queries} & 220 & 73 \\
 & \textbf{Amortized Analysis} & 554 & 70 \\
 & \textbf{Bit Manipulation} & 814 & 90 \\
 & \textbf{Complete Search} & 1886 & 154 \\
 & \textbf{Sorting} & 2234 & 197 \\
 & \textbf{Dynamic Programming} & 2585 & 179 \\
 & \textbf{Greedy Algorithms} & 2649 & 209 \\
 & \textbf{Data Structures} & 4695 & 241 \\
 & \textbf{Unknown} & 14907 & 337 \\
\hline
\end{tabular}
\end{table}

\subsection{Experimental Setup}
The fine-grained labels of the TACO dataset include four dimensions: task theme, algorithm tag, programming skill, and difficulty level. Considering the data volume in the test set and application scenarios, the evaluation of code generation uses two dimensions: programming skill and difficulty level. The distribution of the questions in the TACO dataset across different difficulties and programming skills is shown in Table~\ref{tab:dist}.

Similar to the evaluations of the generation of standard code, for each programming problem, we allow the model to generate 200 attempts, using pass@k (k=1, 10, 100) as the evaluation metric. Therefore, during the model evaluation process, it is necessary to set the hyperparameters of the generation model. Through extensive generation experiments, we found that the model is highly sensitive to the difficulty of the problem. For more challenging problems, the model requires a more diverse range of generated results to potentially produce code that passes the tests. Consequently in the testing process, we use a $top\_p$ setting of 0.95. The temperature settings are differentiated based on problem difficulty levels: we assign a value of 0.2 for easy problems, 0.6 for medium and medium hard problems, and 0.8 for hard and very hard problems.

For general code models, the models are first fine-tuned on the whold training set. Subsequently, on the specified test set, code generation is performed with $1$ to $200$ as random seeds. The generated codes undergo a common truncation scheme as post-processing. The processed code is executed as a program on all test cases, and a program that passes all cases is considered correct code generation.

Since GPT-4 is not a purely code generation model, and considering cost factors, in our evaluation of GPT-4, we use the default settings of top\_p=0.95 and temperature=0.7. We append 'Please write a python program.' to the prompt after the question, generate only once, and extract the Python code block from the markdown as the generated program to calculate pass@1.

\subsection{Results}\label{sec:results}
We selected several classic coding models for evaluation, including codellama-7b, codegen25-mono, starcoder-1b and starcoder-15.5b, and we also conducted tests on GPT-4 for comparison. Evaluations were carried out on 200 questions at five different difficulty levels, and the results are presented in Table~\ref{table:model_performance}:

\begin{table*}[ht]
\centering
\caption{Performances of code generations on TACO coding problems of different difficulty levels}
\label{table:model_performance}
\begin{tabular}{|c|c|c|c|c|c|}
\hline
\textbf{Model Name} & \textbf{Level} & \textbf{Temperature} & \textbf{Pass@1} & \textbf{Pass@10} & \textbf{Pass@100} \\ \hline
\multirow{5}{*}{GPT-4}              & easy           & 0.7                  & 31.50           & -                & -                \\ 
                   & medium         & 0.7                  & 19.00           & -                & -                \\ 
                   & medium\_hard   & 0.7                  & 13.00           & -                & -                \\ 
                   & hard           & 0.7                  & 4.50            & -                & -                \\ 
                   & very\_hard     & 0.7                  & 2.00            & -                & -                \\ \hline
\multirow{5}{*}{codellama-7b-python} & easy           & 0.2                  & 9.32            & 15.15           & 19.99           \\ 
                   & medium         & 0.6                  & 2.38            & 8.28            & 17.51           \\ 
                   & medium\_hard   & 0.6                  & 0.60            & 2.93            & 7.53            \\ 
                   & hard           & 0.8                  & 0.31            & 1.93            & 5.51            \\ 
                   & very\_hard     & 0.8                  & 0.18            & 1.05            & 2.25            \\ \hline
\multirow{5}{*}{codegen25-mono}     & easy           & 0.2                  & 9.06            & 16.12           & 22.52           \\ 
                   & medium         & 0.6                  & 2.42            & 7.83            & 17.23           \\ 
                   & medium\_hard   & 0.6                  & 0.74            & 3.58            & 8.01            \\ 
                   & hard           & 0.8                  & 0.57            & 2.66            & 5.31            \\ 
                   & very\_hard     & 0.8                  & 0.22            & 1.00            & 1.47            \\ \hline
\multirow{5}{*}{starcoder-1b}       & easy           & 0.2                  & 8.95            & 12.01           & 15.51           \\ 
                   & medium         & 0.6                  & 2.22            & 5.32            & 11.16           \\ 
                   & medium\_hard   & 0.6                  & 0.60            & 2.74            & 6.11            \\ 
                   & hard           & 0.8                  & 0.20            & 1.29            & 3.44            \\ 
                   & very\_hard     & 0.8                  & 0.04            & 0.32            & 1.13            \\ \hline
\multirow{5}{*}{starcoder-15.5b}    & easy           & 0.2                  & 11.60           & 18.44           & 23.92           \\ 
                   & medium         & 0.6                  & 3.23            & 8.96            & 19.32           \\ 
                   & medium\_hard   & 0.6                  & 1.43            & 5.43            & 9.41            \\ 
                   & hard           & 0.8                  & 0.58            & 2.79            & 6.39            \\ 
                   & very\_hard     & 0.8                  & 0.18            & 0.85            & 1.75            \\ \hline
\end{tabular}

\end{table*}

The TACO test set poses a significant challenge, with GPT-4 scoring only 31.5 in the pass@1 category for the easy level. Apart from GPT-4, the pass@1 scores of other coding models across all five difficulty levels are generally below 10, and their pass@100 scores are even lower than the pass@1 score of GPT-4.

To further explore the effects on different programming skills, as well as the role of the TACO training set, we conducted experiments on the starcoder-1b model. We used the starcoder-1b model to perform LoRA fine-tuning on the entire training set, resulting in the baseline LoRA model. Additionally, we separately conducted LoRA fine-tuning on subsets of the training set corresponding to different programming skills, resulting in 8 distinct skilled LoRA models. We evaluated both the baseline LoRA and skilled LoRA models, and the evaluation results are presented in the Table~\ref{table:skill_evaluation}:

\begin{table*}[ht]
\caption{Evaluation Results for Different Skill Types}
\label{table:skill_evaluation}
\centering
\begin{tabular}{|c|c|c|c|}
\hline
\textbf{SKILL TYPES} & \textbf{GPT-4} & \textbf{baseline LoRA} & \textbf{skilled LoRAs} \\ \hline
Amortized Analysis   & 11.43 & 0 & 0 \\ \hline
Bit Manipulation     & 10.00 & 1.1 & 1.11 \\ \hline
Complete Search      & 14.29 & 0.32 & 0.32 \\ \hline
Data Structures      & 13.28 & 1.4 & \textbf{1.55} \\ \hline
Dynamic Programming  & 8.94 & 0 & 0 \\ \hline
Greedy Algorithms    & 9.09 & 0 & \textbf{0.36} \\ \hline
Range Queries        & 5.48 & 0 & 0 \\ \hline
Sorting              & 10.66 & 0.7 & \textbf{1.5} \\ \hline
\end{tabular}

\end{table*}

Firstly, GPT-4 demonstrates comparable abilities across various programming skills, with pass@1 scores ranging between 5 and 10, which to some extent indicates the rationality of the skill categorization in the TACO dataset. Secondly, utilizing the TACO training set, which has fine-grained labels, can specifically enhance the performance of code generation models. For example, in Data Structures, Greedy Algorithms, and Sorting, starcoder-1b shows a clear advantage in performance when fine-tuned on specific skills using the TACO training set, compared to being fine-tuned on the entire training set.

\section{Limitations}
During the construction of the TACO dataset, some programming problems were collected from the APPS and CodeContest datasets and were subsequently reclassified. Therefore, after training code models using the TACO dataset, it is not suitable to use APPS and CodeContest as benchmarks to evaluate the code generation capabilities of the models. For assessing programming challenge capabilities, it is recommended to directly use the test set of the TACO dataset.

\section{Other Application Scenarios}
This subsection provides an exhaustive investigation into the versatile utility of the TACO dataset across diverse application domains. The TACO dataset incorporates labels associated with algorithmic skills and is designed to assimilate these skills coherently into code generation frameworks. Specifically, we examine the dataset's prospective contributions to several sectors, including code comprehension, educational pedagogy, algorithmic code recommendation, and applications rooted in Language Learning Models (LLMs).

\begin{itemize}
    \item \textbf{Code understanding:} The TACO dataset is optimally designed for the training of models that are specialized in code interpretation. Within this dataset, each programming challenge is paired with an associated algorithmic competency label. Through the established correlation between the programming challenges problems and their algorithmic labels, the model is empowered to achieve a more nuanced understanding and interpretation of the source code's semantics. As an illustration, the model can autonomously generate annotations that encapsulate pertinent algorithmic metadata within the source code.
    \item \textbf{Education and learning:} In the educational domain, the TACO dataset holds promise for a diverse range of applications, including the creation of programming textbooks, curriculum development, and the assembly of problem repositories for practice. Learners can enhance their programming proficiency by tackling the intricate programming challenges contained in the dataset, each annotated with algorithmic skill labels. These labels facilitate a more profound understanding of the relationship between specific problems and their affiliated algorithms. Moreover, capitalizing on the metadata provided through these labels, educators can streamline the assessment process for student assignments and projects in programming, ensuring adherence to recommended algorithmic methodologies and established best practices.
    \item \textbf{Code Algorithm Recommendations:} The TACO dataset offers promising prospects for contributing to the establishment of an algorithmic recommendation system for coding, which is devised to expedite the process of algorithm selection for software developers. Using the dataset's array of programming challenges and their associated skill labels, developers can efficiently identify algorithmic exemplars that align with the objectives of their ongoing projects, thereby resolving specific coding conundrums. Furthermore, code editing platforms can exploit the metadata embedded in the TACO dataset to furnish developers with more precise suggestions for code auto-completion, encompassing a broad spectrum of algorithmic and data structural options.
    \item \textbf{Potential for application on LLM:} The TACO dataset holds considerable promise for applications grounded in Large Language Models (LLMs). This dataset serves multiple functions: it can expedite the software development process through the automated generation of task-specific code samples, and it enhances the capabilities of LLMs in synthesizing code annotations and documentation to bolster both code comprehensibility and maintainability. Additionally, LLMs that leverage the TACO dataset are capable of addressing queries pertinent to programming and algorithmic topics, providing developers and learners with comprehensive and accurate responses. Collectively, these applications contribute to enhancements in programming efficiency, code quality, and the acquisition of programming knowledge, thereby stimulating ongoing advancements in the fields of programming and algorithms.
\end{itemize}

\bibliographystyle{ijcai24}
\bibliography{ijcai24}
\appendix

\section{More Implementation Details}
\label{appendix:implement}
The principal challenge encountered during the construction of the TACO dataset involved the acquisition of high-quality open-source information. This information predominantly comprised problem descriptions, algorithmic skill labels, Python 3-compliant solutions, and associated test cases from various programming contest platforms. This section elaborates on the particular challenges and informational nuances we encountered during data acquisition across each of these competitive programming platforms.

\begin{itemize}
\item {\textbf{CodeChef:}} During the process of web crawling for problem information from this platform, we discovered that identical content could be represented in multiple HTML syntax formats. To standardize the parsing across these disparate formats, we conducted an extensive series of manual reviews and subsequently developed specialized HTML parsers for each format. These parsers are equipped to manage an array of text formatting elements, such as line breaks and text styles, in addition to converting the HTML source code for mathematical formulas into LaTeX syntax. In the end, we successfully crawled all the problems from the website up to April 2023.

\item {\textbf{CodeForces: }} Given that the APPS and CodeContests datasets already encompass a broad spectrum of problems from this platform, we employed a time-split crawling strategy to maximize efficiency. For problems dating prior to January 2020, we restricted our crawl to the URL, tag, and difficulty level information. This subset of information served two primary purposes: matching with extant problem data and enriching the algorithmic labeling. Conversely, for problems released between January 2020 and April 2023, we conducted a comprehensive data crawl that included problems, solutions, input-output samples, timestamps, time limits, and memory limits. This newly acquired data was then fully integrated into our TACO dataset.

\item {\textbf{HackerRank: }} On the HackerRank platform, we faced multifaceted challenges that included crafting a platform-specific HTML parser and contending with a substantial volume of images in SVG format. These SVG images often depict textual characters and mathematical formulas, which complicates the direct extraction of textual data. To surmount this issue, we employed a two-stages processing approach: initially, SVG vector images were transformed into PNG images featuring black text against a white backdrop; subsequently, the SimpleTex API was harnessed to transcribe the image content into LaTeX syntax format. During our preliminary review of the TACO dataset, we identified approximately 30,000 SVG images within four categories—namely, ``Algorithm'', ``Data Structures'', ``Mathematics'', and ``Python''—that necessitated conversion to LaTeX format on the HackerRank platform. After weighing various factors such as cost and accuracy across different OCR tools (e.g., Mathpix, Pix2Pix, SimpleTex), we selected SimpleTex's API for OCR deployment. It is pertinent to note that errors may occur in SimpleTex's transcriptions, particularly when the SVG images on HackerRank represent overly intricate mathematical formulas. To maintain the dataset's integrity and rigor, we performed manual verifications, revisions, and corrections for the SVG segments in each problem within the HackerRank dataset. We crawled issues from this site that four we needed categories for as of April 2023.

\item {\textbf{Geeksforgeeks: }} GeeksforGeeks serves as a prominent platform for programming education, characterized by meticulously categorized problem content and a well-organized overall structure. However, the website employs various anti-crawling mechanisms, such as request frequency limitations, CAPTCHA validations, and IP address blocks, which presented significant impediments to our data extraction endeavors. Moreover, we encountered frequent parsing errors due to inconsistent formatting and typographical mistakes in the website's problem text. These irregularities hampered our ability to accurately parse the metadata associated with time and space complexity, as stored in the metadata.json file, as well as the input-output samples contained in the input\_output.json file. To navigate these obstacles, we developed a suite of versatile regular expression parsing templates. These templates are engineered to extract maximal problem-related information while maintaining a high degree of accuracy. For data that proved resistant to automated parsing, we conducted manual review and supplementation to ensure maximal fidelity to the original content on the website. We finally managed to crawl the site for 2680 problems through May 2023.

\item {\textbf{CodeWars, Kattis, LeetCode:}} The problems presented on these three educational programming platforms are substantially covered in two pre-existing datasets: APPS and CodeContests. However, these datasets fall short in providing specific algorithmic labels (`tags''), a critical requirement for our research objectives. Consequently, we executed targeted data crawling on each of these websites, with a primary focus on acquiring the algorithmic tags and URL information related to the challenges. We subsequently integrated this newly-acquired information into the existing open-source datasets. This augmentation allows for a systematic algorithmic categorization of these challenges and thereby constitutes a pivotal component of the TACO dataset.
\end{itemize}

While analyzing various programming websites, we observed that certain problems employed images to convey key information, such as graphs or tree structures in a pictorial format. Such information is challenging to accurately represent using LaTeX syntax and could result in incomplete or misinterpreted data when directly inputted into the model. To mitigate this issue, we introduced a new metadata attribute, picture\_num, to the metadata.json file during the data crawling process for the websites CodeChef, CodeForces, HackerRank, and GeeksforGeeks. We tailored the HTML parser for each site to capture this attribute, which records the number of images present in the problem. The objective of incorporating this parameter is to preemptively address potential gaps in information attributable to image-based content during subsequent application phases.

\end{document}